\documentclass[conference]{IEEEtran}
\ifCLASSINFOpdf
  \usepackage[pdftex]{graphicx}
\else
\fi

\usepackage[cmex10]{amsmath}
\usepackage{amssymb}

\usepackage[lined,algonl,boxed]{algorithm2e}

\usepackage{stfloats}

\begin{document}
%
\title{Single-bit-per-weight deep convolutional neural
networks without batch-normalization layers for embedded systems}


\author{
\IEEEauthorblockN{Mark D. McDonnell}
\IEEEauthorblockA{\textit{Computational Learning Systems Laboratory,} \\
\textit{University of South Australia}\\
Mawson Lakes SA, Australia \\
mark.mcdonnell@unisa.edu.au}
\and
\IEEEauthorblockN{ Hesham Mostafa}
\IEEEauthorblockA{\textit{Institute for Neural Computation,} \\
\textit{University of California, San Diego}\\
La Jolla, USA\\}
\and
\IEEEauthorblockN{Runchun Wang and Andr{\'{e}} van Schaik}
\IEEEauthorblockA{\textit{International Centre for}\\\textit{Neuromorphic Systems,} \\
\textit{Western Sydney University}\\
Penrith NSW, Australia\\}
}

\maketitle

\begin{abstract}
Batch-normalization (BN) layers are thought to be an integrally important layer type in today's state-of-the-art deep convolutional neural networks for computer vision tasks such as classification and detection. However, BN layers introduce complexity and computational overheads that are highly undesirable for training and/or inference on low-power custom hardware implementations of real-time embedded vision systems such as UAVs, robots and Internet of Things (IoT) devices. They are also problematic when batch sizes need to be very small during training, and   innovations such as residual connections  introduced more recently than BN layers  could potentially have lessened their impact. In this paper we aim to quantify the benefits BN layers offer in image classification networks, in comparison with alternative choices. In particular, we study networks that use shifted-ReLU layers instead of BN layers. We found, following experiments with wide residual networks applied to the ImageNet, CIFAR 10 and CIFAR 100  image classification datasets, that BN layers do not consistently offer a significant advantage. We found that the accuracy margin offered by BN layers depends on the data set, the network size, and the bit-depth of weights. We conclude that in situations where BN layers are undesirable due to speed, memory or complexity costs, that using shifted-ReLU layers instead should be considered; we found they can offer advantages in all these areas, and often do not impose a significant accuracy cost.
\end{abstract}

\IEEEpeerreviewmaketitle

\section{INTRODUCTION}\label{S:1}

Following its introduction in 2015~\cite{Ioffe.15}, Batch Normalization (BN) layers rapidly became a default layer type in state-of-the-art deep convolutional neural networks (CNNs). In particular for CNNs designed as image classifiers, batch-nomalization is essential for state of the art accuracy on difficult datasets like Imagenet~\cite{He.15a}. However, BN layers have several disadvantages such as:
\begin{itemize}
\item BN layers typically increase the GPU memory requirements and computational load  during training, due to by-default storage of  feature maps for every layer, for use in gradient calculations\footnote{This extra storage can be avoided by recomputing BN layer outputs from its input feature map, but this slows down training slightly.}; additional multiplications needed for BN layers also slows down inference;
\item BN layers and the computation of their parameters are implemented slightly differently in different popular deep learning libraries, resulting in difficulty replicating results and in model portability;
\item BN layers are not well-suited to custom-hardware implementations of training, such as those that use a pipeline approach where all processing needs to be done on one sample independently of all other samples.
\item BN layers create challenges for efficient training when batches are split across multiple GPUs, and as a result sometimes non-exact approximations are used~\cite{Goyal.17};
\item  BN layers can be less effective for models that have to be trained using small batches~\cite{Wu.18}, such as when input images or model sizes are  large enough to fill up GPU RAM;
\item BN  layers can be problematic for datasets with high class-imbalance or when samples in a minibatch are not independent~\cite{Ioffe.17}.
\end{itemize}
The last four disadvantages are due to the fact that BN layers require calculation of batch-wise statistics, namely the mean and variance  of every channel, calculated over all locations in the channel's feature map and all samples in a minibatch, at each point in a network where a BN layer is used. These statistics are less robust for both smaller numbers of samples in a minibatch, and if a batch includes rarely chosen or non-independent samples~\cite{Ioffe.17}.

Some of the potential challenges of BN layers are not as frequently of importance for inference, such as minibatch size. However in the near future, it has been predicted that there will be many  applications where it will be desirable to {\em train} deep neural networks on low-power custom hardware, such as when networks need to be updated frequently using new data, and communication to a data center is too slow or unavailable~\cite{Iandola.16}.
  
For all these reasons, it is desirable to know whether BN layers really are generally necessary for best performance, or whether for particular datasets state-of-the-art accuracy can be achieved without them. However, our main motivation is drawn from seeking to design deep CNN classifiers implemented in custom hardware, that run as fast and efficiently as possible in an embedded system. It is already well-established that convolutional layers can be implemented without use of multipliers, by reducing representation precision of weights or feature maps to 1 bit, hence potentially saving large amounts of chip space and power usage~\cite{Courbariaux.15,Merolla.16,Iandola.16,Rastegari.16,McDonnell.18}. However,  existing investigations in this area often still use BN layers, which if implemented exactly introduce significant complexity, including mandating the use of minibatches, and the need for multipliers, and it is desirable to know whether they can be removed or replaced.

In this paper, we train deep CNNs with and without BN layers, and analyze the resulting accuracy changes.

The paper is structured as follows. In Section~\ref{S2} we review the BN layer definition, and discuss relevant prior research related to the goals of this paper. Then, in Section~\ref{S3}, we describe the CNN architectures  we chose to use for this study, how we vary the use of BN layers in this architecture, and adaptations we need to introduce to enable them to work effectively. Next, Section~\ref{S4} contains our results. Finally, we discuss the implications and significance of our results in Section~\ref{S5}.

\section{Relevant Previous Research}\label{S2}

Depth is crucial for effective learning~\cite{Ba_Caruna14} in  modern neural networks as it allows a network to learn a hierarchical representation of the data by successively composing simpler features into more complex features. Depth, however, poses a number of challenges when training neural networks. The large number of parameters in deep networks typically restricts the optimization methods that could be feasibly used to first order methods like stochastic gradient descent. The loss surface of a deep neural network, however, is highly non-convex and there is no guarantee that starting from an initial parameter point and traversing this loss surface using gradient descent will land in a good local minimum that allows the network to generalize well on unseen data~\cite{Choromanska_etal15,Zhang_etal16a}. A considerable number of techniques and tricks have been developed to allow gradient descent to practically succeed in training deep networks, ranging from random parameter initialization strategies~\cite{Glorot_Bengio10,He.15}, unsupervised pre-training~\cite{Erhan_etal10,Lee_etal09,Ranzato_etal07}, or augmenting gradient descent with gradient history information to more effectively traverse the loss surface~\cite{Kingma_Ba14,Zeiler12}.

The performance of gradient descent is highly dependent on the way in which the optimization problem is parameterized. For example, by re-parameterizing a deep neural network, i.e, by changing the scale and shift of the network's parameters, we can drastically change the curvature of the loss surface~\cite{Dinh_etal17} and the behavior of gradient descent. One of the famous deleterious effects of bad parameterization is the vanishing and exploding gradients problem that plagued early neural networks~\cite{Hochreiter_Schmidhuber97}. To combat such problems, and to yield a more favorable parameterization for gradient descent in general, a broad class of techniques attempt to re-parameterize deep neural networks through various forms of normalization. Such normalization techniques operate on various quantities in the network, such as activations or weights.

Perhaps the most popular normalization technique is batch-normalization~\cite{Ioffe.15}, which is described in more detail in the next subsection. Batch-normalization not only accelerates gradient descent learning, it also has a beneficial regularization effect~\cite{Morcos_etal18}. Batch-normalization operates on pre-activations (the inputs to neurons before the activation function is applied); it rescales and shifts the pre-activations of each single neuron so that they have zero mean and unity variance across the mini-batch samples, in each channel. During training, the network's response to an example  thus depends on the other examples that accompany it in a mini-batch. This dependence, however, could prove undesirable in some situations~\cite{Salimans_etal16}. Batch-normalization requires the use of a mini-batch that is large enough to yield reliable pre-activation statistics. Increasing mini-batch size, however, often leads to a decrease in the network's generalization performance~\cite{Keskar_etal16}.

A related normalization technique that does not depend on mini-batch statistics is layer normalization~\cite{Ba_etal16}. Layer normalization normalizes the pre-activations of neurons in a layer so that they have zero mean and unity variance. These statistics are calculated across all pre-activations in a layer, unlike batch-normalization which calculates them for each neuron individually across the mini-batch. One downside, however, is that different neurons, especially in convolutional layers, can have widely different input statistics, so normalizing all of them using the same coefficients is poorly motivated. Instead of normalizing the activations, ref~\cite{Salimans_etal16a} normalizes by the norm of the input weight vector of each neuron. An extra parameter is introduced for each weight vector to explicitly control its length. The performance of weight normalization closely matches that of batch-normalization while avoiding its major downside: the dependence on mini-batch statistics. Another innovation introduced by~\cite{Salimans_etal16a} is the ``mean-only BN'' layer, in which layer inputs are centered according to the mean of a batch, in conjunction with weight normalization.

Recently, yet another method was introduced and shown to outperform BN: group normalization~\cite{Wu.18} while avoiding some of its downsides.

The choice of activation function has a large impact on the performance of gradient descent or backpropagation. When errors are backpropagated through a layer, they are scaled by the derivative of the activation function of the neurons in that layer.  If these derivatives are small, then errors are effectively blocked from propagating backwards. Saturating activation functions such as the logistic sigmoid, or hyperbolic tan, are particularly vulnerable to this effect as their derivatives are very small when their input is far from zero~\cite{Lecun_etal98}. The use of non-saturating activation functions such as Rectified Linear Units (ReLUs) has partially alleviated this problem. A ReLU activation is zero for negative inputs and the identity for positive inputs. ReLUs, however, cannot have an output that is zero mean across the training examples as they do not produce negative outputs. Exponential Linear Units (ELUs)~\cite{Clevert.15} address this issue by having a saturating negative output if the input is less than zero. ELUs have been shown to perform extremely well without any form of explicit normalization suggesting that they intrinsically normalize the levels of activation in the network.    

However, shifted Rectified Linear Units (sReLUs) offer nearly all the same benefits as ELUs, but without the need to calculate exponentials, and given our motivation of minimizing computational load, they are our focus here instead of ELUs. Indeed, our results in Figures 5 and 6 show that networks using sReLU and ELU activations do not have significant differences in accuracy. 

For similar reasons, we aim to establish if we can avoid alternatives to BN like group normalization~\cite{Wu.18}, where  complexity and additional computation is introduced for computing statistics and carrying out normalizations by non-constant factors.

\subsection{Review of BN layers} 

Batch-normalization (BN) layers have been shown to help deep neural networks produce better accuracy following training. The usual explanation for this is that they help ``reduce internal covariate shift''~\cite{Ioffe.15}. This view has been recently challenged~\cite{Santurkar.18}. However it is clear that the use of BN layers enables higher learning rates, and hence faster convergence during training, and diminishes the importance of good initial conditions for convolutional layers~\cite{Ioffe.15,Ioffe.17}. 

BN layers are applied to a minibatch of feature maps, which typically can be represented as a 4-axis tensor, ${\bf x}\in \mathcal{R}^{K,H,W,C}$, where $K$ is the size of the minibatch, $H$ and $W$ are the height and width of the feature maps, and $C$ is the number of channels. BN layers operate on a per-channel basis, with channel-specific gain and shift parameters, and make use of channel-specific calculations of the mean and variance of the inputs to the channel in a minibatch. Mathematically, if $x_{k,h,w,c}$ is the value of the feature map at location $h,w$ in the $c$-th channel of the $k$-th sample, then BN layers transform that value according to
\begin{equation}
\hat{x}_{k,h,w,c} = g_c\left(\frac{x_{k,h,w,x}-\mu_c}{\sqrt{\sigma^2_c+\epsilon}}\right)+o_c,
\end{equation}
where $\mu_c$ and $\sigma^2_c$ are calculated as
\begin{equation}
\mu_c = \frac{1}{KHW}\sum_k\sum_h\sum_wx_{k,h,w,c}
\end{equation}
 and
 \begin{equation}
    \sigma_c^2 = \frac{1}{KHW}\sum_k\sum_h\sum_w (x_{k,h,w,c} -\mu_c)^2.
 \end{equation}
 The remaining parameters are:
 \begin{itemize}
     \item gain, $g_c$, and shift $o_c$, which are usually learned in the same manner as weights in convolutional layers;
     \item $\epsilon$, which ensures division by zero cannot happen in the event of zero variance, and can also act like a regularizer; it is not widely appreciated that accuracy can depend on the exact choice of $\epsilon$ and that inference performance is sensitive to inadvertent changes in $\epsilon$ compared to training. It is also noteworthy that some popular deep learning libraries define $\epsilon$ differently, by taking it outside the square root, which changes subtely but sometimes significantly the performance of otherwise identical models. 
 \end{itemize}
 
 There are two major differences between BN layers and other layers:
 \begin{enumerate}
     \item calculation of layer outputs depend on all samples in a minibatch and cannot be computed independently for each sample;
     \item During training, minibatch  means and variances are calculated. For inference, each mean and variance is an additional parameter that forms part of the trained model, but unlike most parameters derived from training data, these are not learned, but rather are computed.  Most popular libraries compute these values during training by calculating exponential moving averages over batches as training progresses. We have found that this can sometimes give misleading and sub-par performance during monitoring on a validation set as training progresses, because parameters change over the window in which averages are created. Instead, we favour calculation of batch means and variances using multiple training batches while training is frozen, as described in Algorithm 2 in the original BN paper~\cite{Ioffe.15}.
 \end{enumerate}
 
\subsection{BN layer scale and shift can be detrimental} 
 
Recent work demonstrated the surprising result that not learning BN scale and shift parameters can actually be beneficial, leading to reduced error rates for CIFAR 10 and CIFAR 100 in a wide residual network~\cite{McDonnell.18}. This result was found to be the case for both full-precision networks, and for ``1-bit-per-weight'' versions of the same networks.

\subsection{Shifted Rectified Linear Unit and Exponential Linear Unit}

\begin{figure}[h]
\begin{center}
{\includegraphics[scale=0.35]{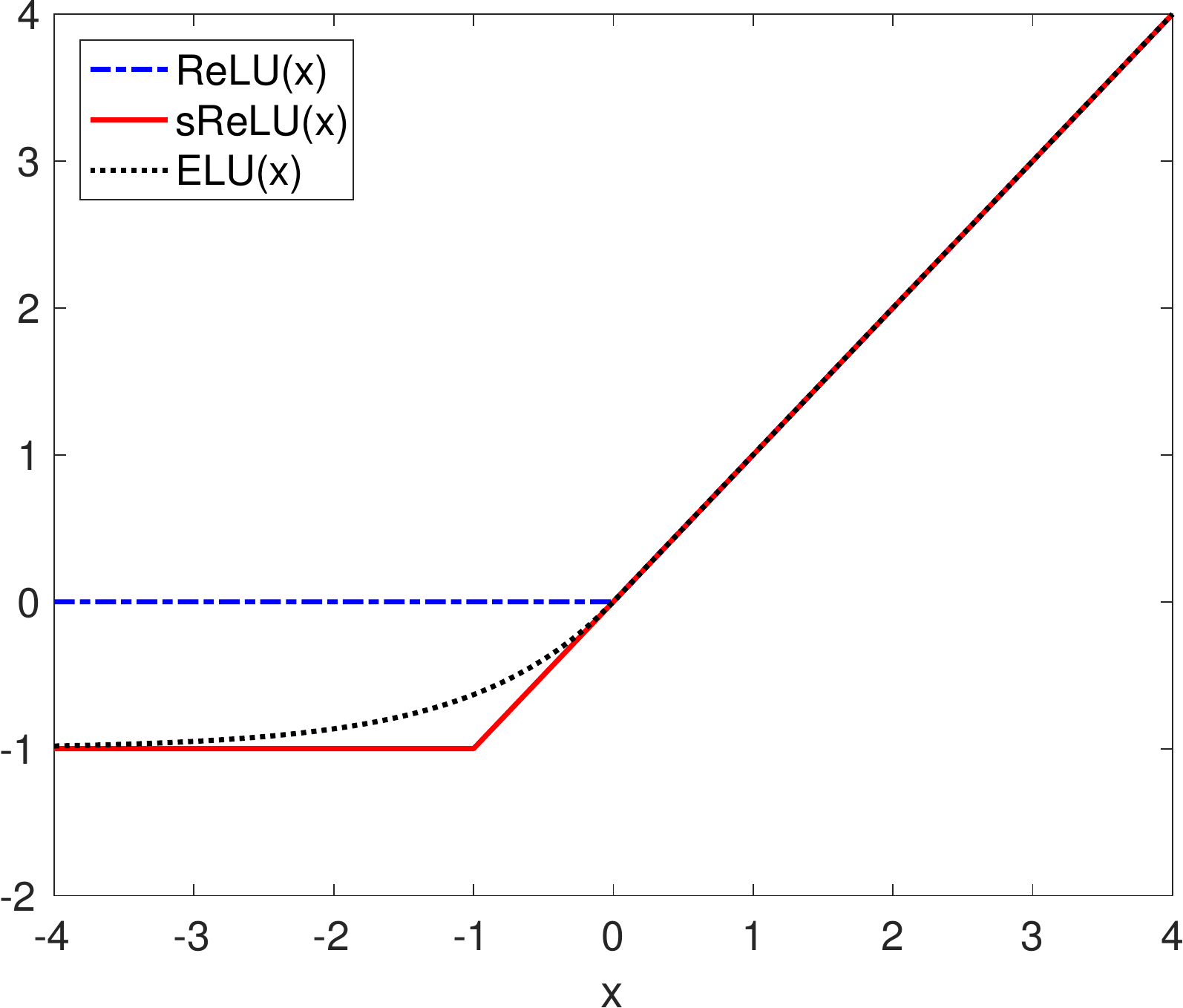}}
\end{center}
\caption{{\bf  Shifted Rectified Linear Unit (sReLU) activation function.} The sReLU activation function lets negative inputs pass through, between 0 and some negative constant, in this case equal to $-1$. While the Exponential Linear Unit (ELU) is more popular, we have found sReLU to be equally effective, and less computationally demanding, due to avoiding calculation of an exponential.}\label{fig:srelu}
\end{figure}

It has already been shown that removal of BN layers can somewhat be compensated for by using exponential linear units (ELUs)~\cite{Clevert.15} instead of the combination of BN and ReLU. The mathematical definition of these two activation functions are
\begin{equation}
    {\rm ELU}(x) = x\mathcal{I}(x)+(\exp(x)-1)(1-\mathcal{I}(x)),
\end{equation}
where $\mathcal{I}(x)$ is the Heaviside step function,
and
\begin{equation}
    {\rm sReLU}(x) = {\rm max}(-1,x).
\end{equation}
Both can be generalized to a different minimum output value, but we consider only the case of $-1$. See Figure~\ref{fig:srelu} for plots of the sReLU and ELU characteristics, in comparison with ReLU.

When ELUs were introduced, it was shown that they outperform shifted rectified linear units (sReLUs)~\cite{Clevert.15}. However, in our experiments with more recently advanced networks and training methods not available at the time of~\cite{Clevert.15}, we did not observe any significant difference between ELUs and sReLUs. Moreover, ELUs are computationally expensive compared with sReLU, both because sReLUs do not require calculations of exponentials in the forward pass, nor storage of their values for use in the backward pass during training. We show here that shifted ReLUs are an effective and efficient alternative to ELUs, and seek to quantify how much accuracy penalty is incurred by replacing BN layers.

\section{Methods}\label{S3}

\subsection{Baseline network architecture and training}

\begin{figure*}[ht]
\begin{center}
{\includegraphics[scale=0.7]{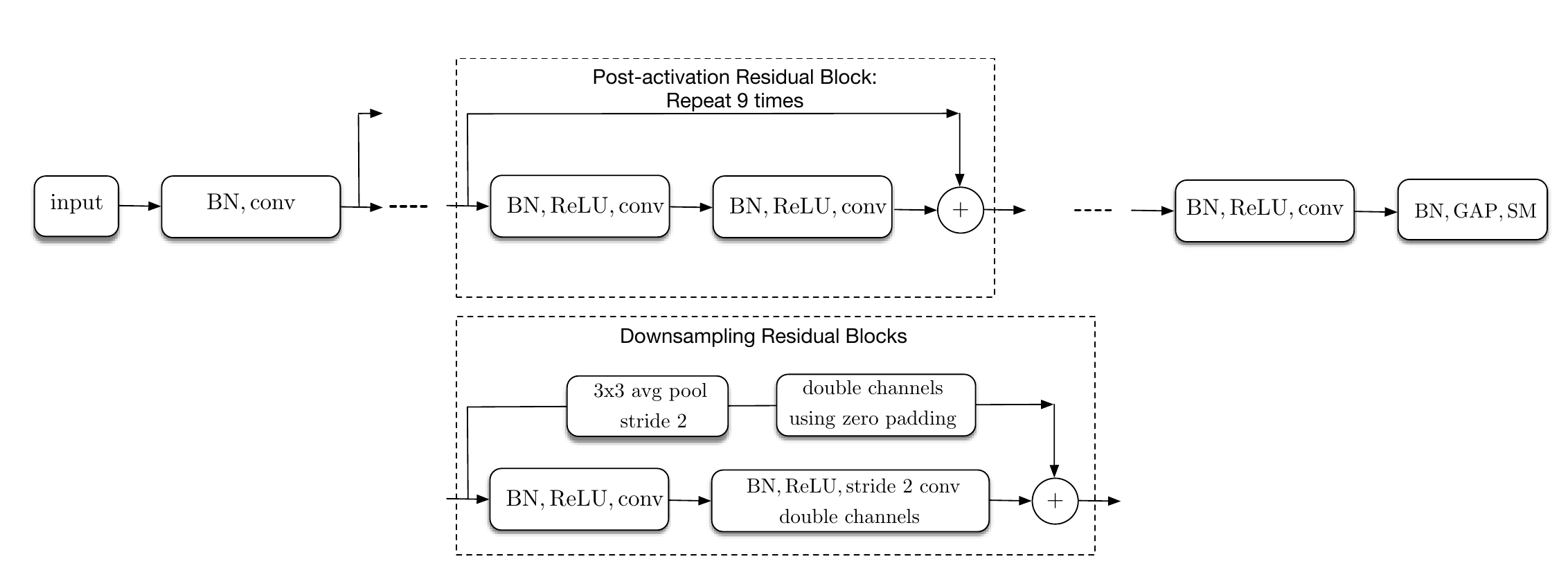}}
\end{center}
\caption{{\bf  Wide ResNet architecture for Baseline CIFAR models where BN layers are used.} This architecture is nearly identical to that of~\cite{McDonnell.18}, except here there is no optional ReLU applied to the input. Note the ordering of the final layers, where global average pooling (GAP) is used after a final 1$\times$1 convolutional layer, that reduces the number of channels to equal the number of classes, and then feeds directly to the softmax output (SM).}\label{fig:arch1}
\end{figure*}

Our experiments are based on wide residual networks~\cite{Zagoruyko.16} (see Figure~\ref{fig:arch1}) that use  the post-activation architecture~\cite{He.16}. Using the nomenclature of~\cite{Zagoruyko.16}, for CIFAR 10 and CIFAR 100 we used depth 20 and widths of 4$\times$ and 10$\times$ and for ImageNet depth 18 and widths 1$\times$ and 2.5$\times$,  meaning in both cases that the first layer of convolutional weights had either 64  or 160 output channels. The design we used has a few differences to~\cite{Zagoruyko.16}, such as that the final weights layer is a $1\times 1$ convolutional layer applied before the global average pooling layer---see Figure~\ref{fig:arch1} which illustrates this aspect, and the overall design. More details are described in~\cite{McDonnell.18}. Note also that in the design of~\cite{McDonnell.18} a BN layer was applied to the RGB input channels prior to  the first convolution layer. Here we do the same for all variations, including the sReLU one. Provided that the shift and scale factors are not learned, this BN layer is equivalent to preprocessing raw data, and does not need to be considered to form a network layer.

Training was carried out similarly to~\cite{McDonnell.18}, i.e.~we used backpropagation and stochastic gradient descent, with minibatches of size 125 samples, momentum of 0.9 and weight decay with a value of 0.0005. No biases are used. Weights were initialized using the method of~\cite{He.15}, while BN gains were initialized to 1 and shifts to 0. No convolutional layer biases were used. Unlike~\cite{McDonnell.18}, we did not use a warm restart learning rate schedule, as we found this could sometimes lead to divergence following a restart with the sReLU networks. However, we did use a cosine learning-rate decay schedule, starting at an initial value of $0.1$ and finishing at $10^{-5}$ after 300 epochs (CIFAR) or 60 epochs (ImageNet). For networks where BN layers were used, we computed the mean and variance statistics following the end of training, by calculating averages over all minibatches in 1 epoch, with learning turned off.

During training, each image selected for a minibatch was augmented using standard methods as in~\cite{McDonnell.18}.  In addition, cutout augmentation~\cite{Devries.17} was used for CIFAR 10 and CIFAR 100, with the same design as in~\cite{McDonnell.18},  with a patch size of 18 pixels.

\subsection{1-bit-per-weight networks}

\begin{figure}[h]
\begin{center}
{\includegraphics[scale=0.6]{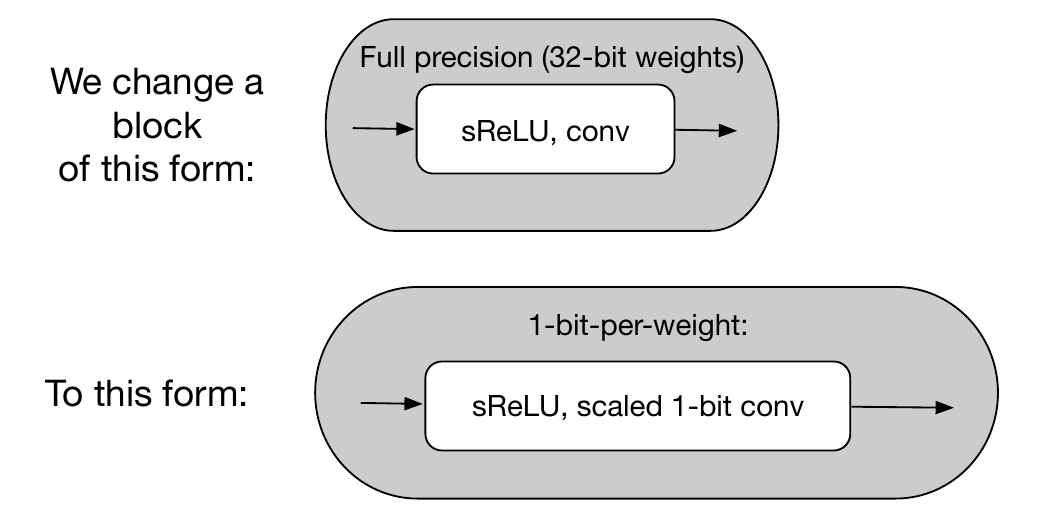}}
\end{center}
\caption{{\bf  Changes when training for 1-bit-per-weight.} When we train 1-bit-per-weight networks following the method of~\cite{McDonnell.18}, we apply the sign operator to full-precision copies of weights during training, and then scale by a constant equal to the initial standard deviation of the weights according to the method of~\cite{He.15}.}\label{fig:birn}
\end{figure}

As well as training networks with the usual 32 bit floating point precision for all variables, including learned weights, we also trained networks using a method for enabling storage of learned weights and inference to take place using 1-bit values. We followed the method of~\cite{McDonnell.18}; differences in training are summarised for the case of sReLUs in Figure~\ref{fig:birn}.

We emphasize that the main motivation of this paper is to examine whether methods for enabling reduced-precision representations in deep neural networks, such as~\cite{McDonnell.18}, still work effectively when batch-normalization layers are removed.

\subsection{Shifted ReLUs}

\begin{figure*}[ht]
\begin{center}
{\includegraphics[scale=0.7]{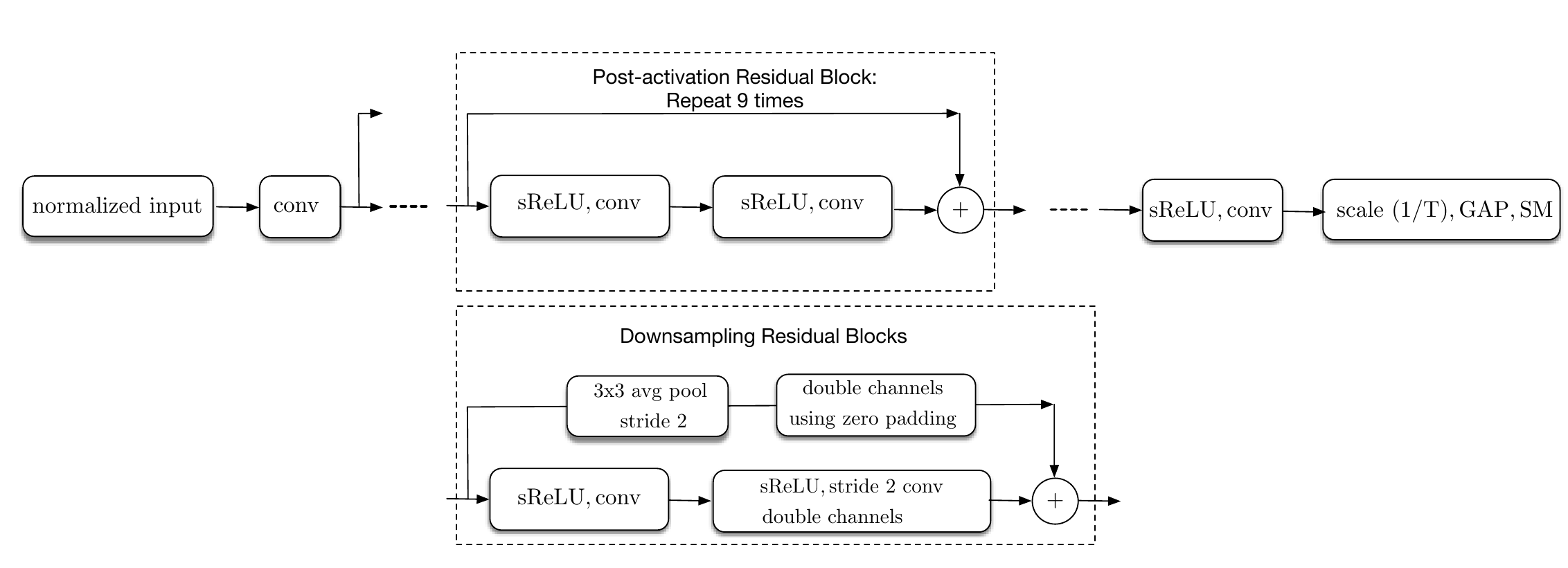}}
\end{center}
\caption{{\bf  Wide ResNet architecture for CIFAR when all BN layers are replaced by sReLUs.} The architecture is identical to that of Figure~\ref{fig:arch1} except that (i) all BNs have been removed and ReLUs have been replaced by shifted ReLUs (sRELU); (ii) a scale layer (multiplies all inputs by a constant) has been inserted before the global average pooling (GAP) layer; and (iii) the input is now normalized using pre-preprocessing applied to each of the RGB channels.}\label{fig:arch2}
\end{figure*}

We conducted experiments using the same architecture as in Fig.~\ref{fig:arch1}, but where  (i) all BN layers except the final one closest to the output are replaced by shifted ReLUs and (ii) where all BN layers are replaced by shifted ReLUs. The architecture for the latter case is shown in Figure~\ref{fig:arch2}.

\subsection{Training innovations for networks with BN layers removed}

For the model where all BN layers but the final one were removed, we found we did not need to change any aspect of training relative to our baseline models.

However, without any BN layers, we found that training diverged, or converged very slowly. A partial remedy to this problem is to simply reduce the learning rate, a fact consistent with one of the advantages of using BN layers, i.e.~that the initial learning rate can be set higher~\cite{Ioffe.15} than in networks without BNs.  However, this leads to slow convergence and increased error rates.

We investigated why this was the case. We found, surprisingly, that using just a single BN layer between the final convolutional layer and the global average pooling layer  was sufficient to enable the network to be trained exactly like the baseline models. 

Based on this observation, we hypothesized that the main reason that networks with BN layers enable a larger learning rate is due to the standarized scaling the final BN layer imparts on gradients from the output prior to backpropagation to weight layers. Indeed, we observed that without a final BN layer, the distribution of gradients at the output early in training have a longer tail then when BN is included.

We therefore introduced in our shifted ReLU networks a constant scaling layer to replace the BN layer between the final convolutional layer and global average pooling layer. For simplicity we have set the constant scaling identically for all channels (which in our design at this point in the network is equal to the number of classes). 

Due to the linearity of the global average pooling layer, changing this scaling is equivalent to changing the temperature in the softmax layer, from its default value of 1. That is, our approach corresponds to employing a final softmax layer of the form
\begin{equation}
{\rm SM}_i(x) := \frac{\exp{\left(\frac{x_i}{T}\right)}}{\sum_{j=1}^N \exp{\left(\frac{x_j}{T}\right)}},
\end{equation}
where $T$ is the temperature---see, e.g.~\cite{Hinton.15,Guo.17}.

We found a value for the temperature between approximately 30 and 100 to enable training to take place identically to the baseline models, including the same high initial learning rate. Lower values of $T$ tended to result in failure to converge shortly after training commenced.

Note that although the gradient propagated back is linearly scaled by $1/T$, changing from the default of $T=1$ is not equivalent to simply changing the learning rate by a factor of $T$. This is due to the nonlinearity in the softmax layer. Increasing $T$ has the effect of moving softmax outputs away from 0 or 1, thereby increasing the entropy of the output vector. In turn, this means gradients propagating backwards early in training have a distribution with lower standard deviation.

That this temperature scaling is all that is needed to ensure a high learning rate suggests that the main problem with larger learning rate for models without BN layers is simply that early in training, the gradients calculated at the output of the network are too high for high learning rates, and that the normalization of the final BN layer compensates for this.





\section{Experiments and Results}\label{S4}

We trained the following variations of our wide residual networks on both CIFAR 10 and CIFAR 100, all for both width 4 and width 10. 
\begin{enumerate}
\item  Baseline 1: networks as in Figure~\ref{fig:arch1},  with conventional BN layers where scales and shifts are learned.
\item Baseline 2: the same as Baseline 1, but without any scales and shifts learned, as in~\cite{McDonnell.18}.
\item A single BN layer at the end of the networks only (no scale and shift learned), with all other BN-ReLU combinations replaced by sReLU.
\item No BN layers -- all replaced by sReLU, as in Figure~\ref{fig:arch2}.
\item No BN layers -- all replaced by ELU~\cite{Clevert.15}.
\item No BN layers -- all replaced by sReLU, except for a `mean-only BN' layer at the end of the network~\cite{Salimans_etal16a}, without a learned bias.
\end{enumerate}

\begin{figure}[h]
\begin{center}
{\includegraphics[scale=0.45]{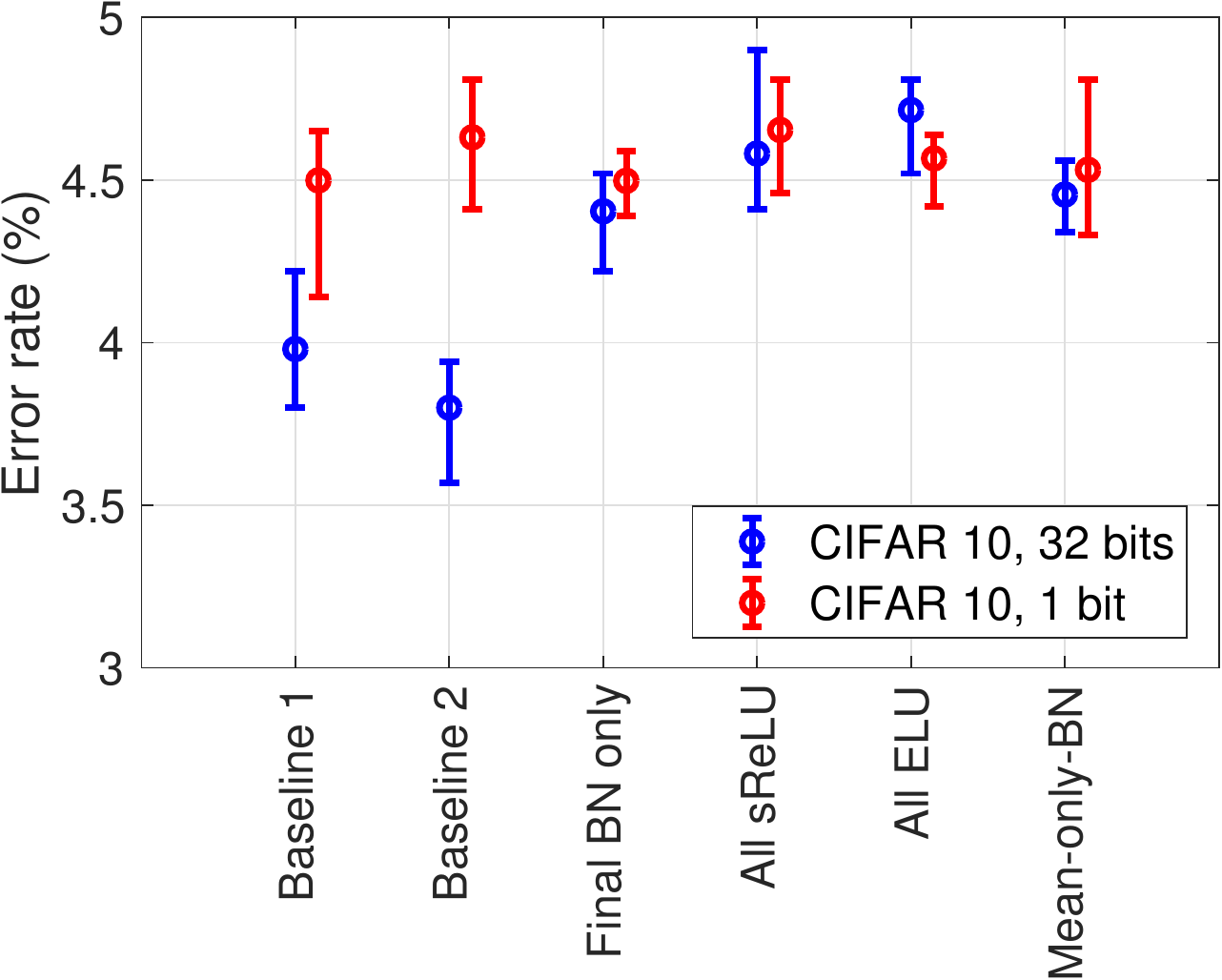}}
\end{center}
\caption{{\bf  Spread of results: CIFAR 10, Width 4.} The circle markers show the mean from 10 repeated runs for each of the 4 model types, using different random seeds for each repeat, but the same seed for each model. The error bars indicate the maximum and minimum errors over the 10 repeated runs.}\label{fig:res1}
\end{figure}

\begin{figure}[h]
\begin{center}
{\includegraphics[scale=0.45]{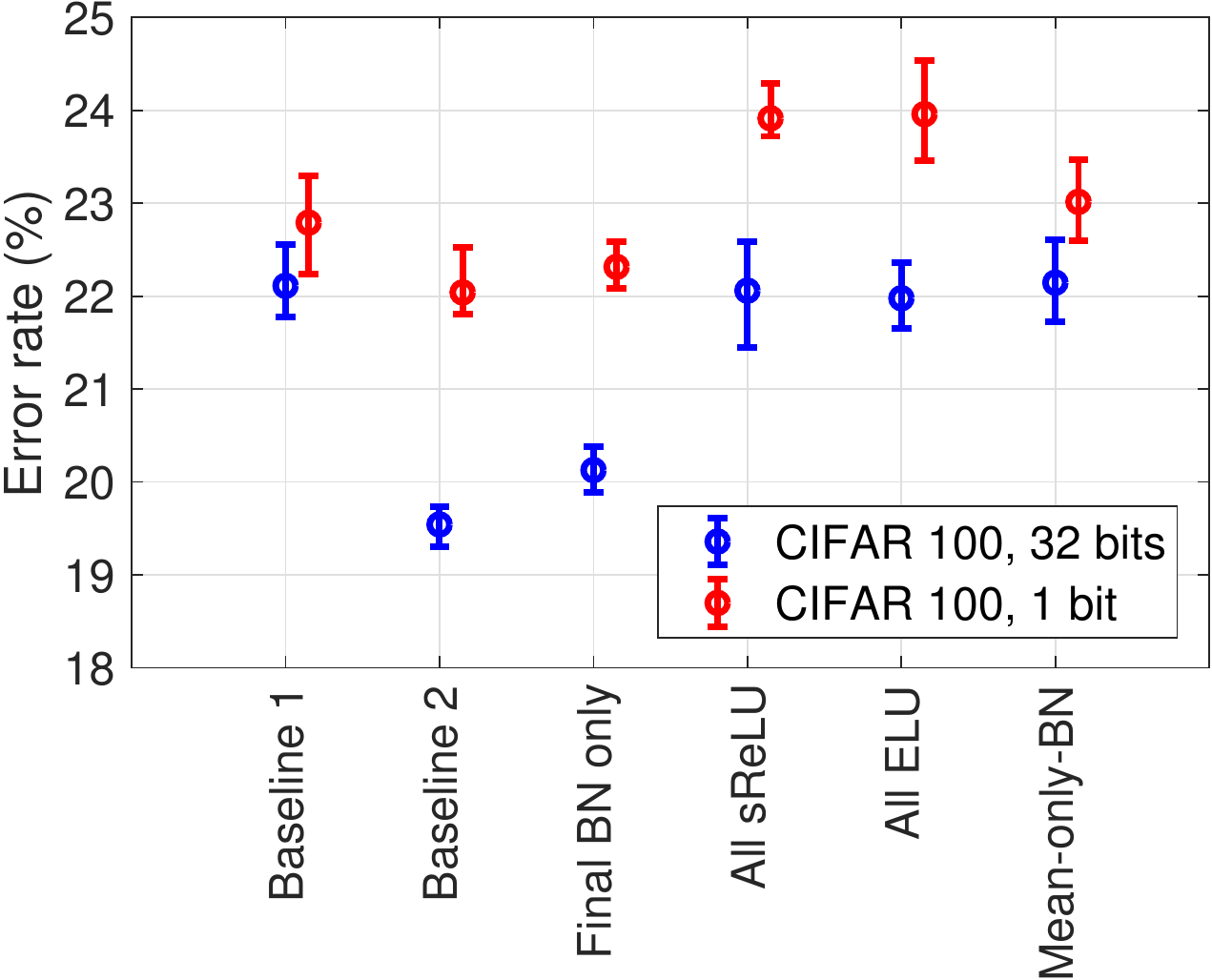}}
\end{center}
\caption{{\bf  Spread of results: CIFAR 100, Width 4.} The circle markers show the mean from 10 repeated runs for each of the 4 model types, using different random seeds for each repeat, but the same seed for each model. The error bars indicate the maximum and minimum errors over the 10 repeated runs.}\label{fig:res2}
\end{figure}

Our results for width-4 and width-10 ResNets are summarized in Tables I and II, while Figures~\ref{fig:res1} and~\ref{fig:res2} show the mean and spread of accuracies for width-4 ResNets for CIFAR 10 and CIFAR 100 over 10 runs. 

For ImageNet, we compared width-1 and width-2.5 networks, and an ensemble of 3 width-1 networks, for the case of Baseline 1, and case 3 in the above list. The results are summarised in Table III.

Our conclusions drawn from the results are left for Section~\ref{S5}.

\begin{table*}[h]
\caption{{\bf CIFAR 10: Test-set error-rates.} The entries for ResNet 20-4 networks are the mean value from 10 repeats. The entries for ResNet 20-10 networks are for single runs. The final column shows the difference between the errors for sReLU only networks compared with the best result in each row, as indicated in bold font.} \label{Table00}
\begin{center}
\footnotesize
\begin{tabular}{|c|c|c|c|c|c|c}
\hline
{\bf Model} &{\bf Baseline 1} & {\bf Baseline 2} & {\bf Final BN only} & {\bf sReLU only} &{\bf sReLU gap}\\
\hline
ResNet 20-4, 32 bit weights & {3.97}\%& {\bf 3.80}\% &4.42 \% & 4.67\% & 0.87\%\\
\hline
ResNet 20-10, 32 bit weights & {\bf 3.29}\%&3.57\% &3.79\% & 4.36\%& 1.07\%\\
\hline
\hline
ResNet 20-4, 1 bit weights & {\ 4.51}\%& 4.65\%&{\bf 4.47}\% &4.66\% & 0.19\%\\
\hline
ResNet 20-10, 1 bit weights &3.83\% & {\bf 3.65}\%&4.00\% & 3.74\%& 0.09\%\\
\hline
\end{tabular}
\end{center}
\end{table*}%


\begin{table*}[h]
\caption{{\bf CIFAR 100: Test-set error-rates.} The entries for ResNet 20-4 networks are the mean value from 10 repeats. The entries for ResNet 20-10 networks are for single runs. The final column shows the difference between the errors for sReLU only networks compared with the best result in each row, as indicated in bold font.}\label{Table01}
\begin{center}
\footnotesize
\begin{tabular}{|c|c|c|c|c|c|c}
\hline
{\bf Model} &{\bf Baseline 1} & {\bf Baseline 2} & {\bf Final BN only} & {\bf sReLU only}& {\bf sReLU gap}\\
\hline
ResNet 20-4, 32 bit weights & 22.10\%& {\bf 19.53}\% &20.18\% & 22.24\%& 2.69\%\\
\hline
ResNet 20-10, 32 bit weights & 20.99\%&{\bf 17.05}\% &18.25\% & 20.97\%& 3.92\%\\
\hline
\hline
ResNet 20-4, 1 bit weights & 22.82\%& {\bf 22.07}\%&{ 22.31}\% &23.69\% & 1.60\%\\
\hline
ResNet 20-10, 1 bit weights &20.61\% & {\bf 18.22}\%&19.42\% & 20.74\%& 2.52\%\\
\hline
\end{tabular}
\end{center}
\end{table*}%


\begin{table*}[h]
\caption{{\bf ImageNet: Validation-set top-5 error-rates.} Centre crop means a single $224\times224$ crop from the centre of the validation image was used. Multicrop means 25 different crops were run used and their predictions averaged before classifying, similar to~\cite{He.16}.
}\label{Table02}
\begin{center}
\footnotesize
\begin{tabular}{|c|c|c|c|c|c|c|}
\hline
{\bf Bits per weight} &{\bf Model} &{\bf \# Learned parameters} & {\bf Test mode} &  {\bf Baseline 1}  & {\bf Final BN only} &{\bf  Gap}\\
\hline
32& ResNet 18-1&11.5M& centre crop & 12.41\% & 15.50\% & 3.01\%\\
\hline
32& ResNet 18-1&11.5M& multi crop &  9.03\% & 14.70\% & 5.67\%\\
\hline
32& Ensemble of 3 ResNet 18-1&34.5M &centre crop & 10.70\% & 14.00\% & 3.30\%\\
\hline
32& Ensemble of 3 ResNet 18-1&34.5M& multi crop & 7.95\% & 12.30\% & 4.35\%\\
\hline
32& ResNet 18-2.5 &70M& centre crop & {{9.20\%}} & 9.51\% & 0.3\%\\
\hline
32& ResNet 18-2.5 &70M& multi crop &  {{6.91\%}} & 8.83\%& 1.92\%\\
\hline
\hline
1& ResNet 18-1&11.5M& centre crop & 17.55\%&23.66\% &6.11\%\\
\hline
1&ResNet 18-1&11.5M& multi crop &12.80\% &19.94\%&7.14\%\\
\hline
1&Ensemble of 3 ResNet 18-1&34.5M &centre crop &15.48\% &22.18\% &6.70\%\\
\hline
1&Ensemble of 3 ResNet 18-1&34.5M& multi crop &11.38\% &18.85\%&7.47\%\\
\hline
1&ResNet 18-2.5&70M& centre crop & {{11.51\%}}  & 11.81\%&0.3\% \\
\hline
1&ResNet 18-2.5 &70M&  multi crop &  {{8.48\%}}  &10.03\% &1.58\% \\
\hline
\end{tabular}
\end{center}
\end{table*}%

\section{Discussion}\label{S5}

\subsection{The impact of removing BN layers}

Our results indicate the following:
\begin{itemize}
    \item The importance of BN layers is data-set and model-size dependent. Our findings show that the accuracy loss when replacing all BN layers with shifted ReLUs is larger on CIFAR 100 than on CIFAR 10. For all three datasets, and especially ImageNet,  the accuracy loss is very large for width-1 (non-wide), but not for wide variants, indicating BN is much more important for smaller models.
    \item For CIFAR, there is no clear advantage in using ELU layers rather than sReLU.
    \item The impact of removing BN is larger for full-precision weights compared with 1-bit-per weight.
    \item Baseline 1, where gains and shifts are learned is about as  good for CIFAR 10 as Baseline 2 where they are not, but Baseline 2 is clearly better for CIFAR 100. 
    \item Using a single BN-layer at the end of networks was mostly markedly better than using all sReLUs, and in some cases nearly as good as the best baseline.
    \item For width 4 networks, the variations within models across runs for CIFAR 10 and 1-bit-per-weight indicated that any model could produce best results. For CIFAR 100, the first three models exhibited this effect, but the sReLU network clearly had a small gap in performance compared to other models. For 32-bit-per-weight models, gaps were larger, but the top two models for CIFAR 10 were the two baselines, while for CIFAR 100 they were Baseline 2 and the final BN only model.
    \item The mean-only-BN model outperforms the All ReLU and All ELU networks for CIFAR 100 and 1-bit per weight, halving the error rate gap to Baseline 2.
\item Consistent with~\cite{McDonnell.18}, for CIFAR 100, Baseline 1 where scales and shifts are learned causes a dramatic drop in accuracy relative to Baseline 2. Unlike~\cite{McDonnell.18}, this is no longer the case for CIFAR 10, which might be because here we do not use a ReLU applied to the input.
\end{itemize}
From these observations, we propose the following:
\begin{itemize}
    \item {\bf Conclusion 1}: Removing all BN layers can be expected to cause an accuracy penalty, but this penalty is potentially small, depending on the dataset and network architecture.
    \item {\bf Conclusion 2}: Removing all but the final BN layer is a potentially viable option for getting the benefits of removing most BN layers, but without as much accuracy loss as removing all.
    \item {\bf Conclusion 3}: There is no consistently best way to design networks with BNs; in some cases learning scales and shifts is beneficial, but in other cases it causes a big drop in performance.
     \item {\bf Conclusion 4}: There is less impact on accuracy in the case of 1-bit-per-weight compared with full precision weights. Indeed, for CIFAR 10, sReLU networks and 1-bit-per-weight had an accuracy gap no larger than 0.2\%, and only 0.3\% for width-2.5 ImageNet with centre cropping, suggesting sReLU as a viable option when 1-bit-per-weight is used.
\end{itemize}

\subsection{Significance for custom hardware implementations}

Hardware acceleration of CNNs is necessary to provide real-time embedded vision systems, e.g., UAVs and Internet of Things (IoT) devices, because existing systems using CPUs are too slow. To increase speed, most software-based CNNs use GPUs. However, it is difficult to deploy GPUs in embedded systems, since they consume a significant amount of power. Thus, custom rather than general-purpose hardware-based CNNs are desired for low-power and real-time embedded vision systems. 
 
In hardware-based acceleration systems, most power consumption comes from the computation modules, e.g., the multipliers and adders, and from  accessing of the data, particularly the weights. By using binary weights, the multiplier can be replaced by a multiplexer, which consumes orders of magnitude less power. In modern CMOS technologies, e.g.,~28nm, a binary convolutional operation with multiplexers achieves a power efficiency up to 230 1b-TOPS/W~\cite{Moons.18}. More importantly, using binary weights enables use of only on-chip memories, e.g., SRAMs, to store these weights. Accessing data stored in external memories would consume an order of magnitude more power (10$\times$) than the computation itself~\cite{Chen.16}. 
 
The use of batch-normalization layers might maintain maximum classification accuracy but at the cost of extra silicon area and  computation and thus more power consumption. Particularly, it will require significant amount of silicon area to implement the nonlinear square and square-root operations in the conventional batch-normalization. What makes it worse is that these operations impede low bit-width quantization techniques~\cite{Ando.17}. One can easily implement the shifted ReLU activation function with a tiny silicon area, while achieving comparable accuracies to networks with BN layers.

\subsection{Future work}

In future, it will be valuable to try to devise new low-complexity methods that narrow the accuracy gap between networks that use BN layers, and the ones outlined here.

\section*{ACKNOWLEDGMENT}

This work was supported by a Discovery Project funded by the Australian Research Council (project number DP170104600).


\end{document}